\pdfoutput=1
\documentclass[11pt]{article}

\usepackage[margin=0.95in]{geometry}
\usepackage[T1]{fontenc}
\usepackage[utf8]{inputenc}
\usepackage{lmodern}
\usepackage{microtype}
\usepackage{setspace}
\usepackage{amsmath,amssymb,mathtools}
\usepackage{booktabs}
\usepackage{multirow}
\usepackage{float}
\usepackage{graphicx}
\usepackage{xcolor}
\usepackage{xurl}
\usepackage[numbers,sort&compress]{natbib}
\usepackage[colorlinks=true,linkcolor=blue!60!black,citecolor=blue!60!black,urlcolor=blue!60!black]{hyperref}

\newfloat{algorithm}{htbp}{loa}
\floatname{algorithm}{Algorithm}

\title{\vspace{-1.4em}\Large\textbf{Ontology-Compiled 2D Clinical State Fields with Bidirectional Diffusion Encoders for Zero-Training Cross-Ontology Medical Coding}}

\author{
\normalsize \textbf{Zeqiang Wang}\\[0.25em]
\small Department of Computer Science, University of Surrey, Guildford, United Kingdom\\
\small \texttt{zeqiang.wang@surrey.ac.uk}
}
\date{}

\begin{document}
\maketitle
\vspace{-1.0em}
\begin{abstract}
\noindent\textbf{Objective:} Healthcare systems continually revise clinical taxonomies, update reimbursement schedules, and harmonize cross-border registries. Existing automated coding models bind clinical text understanding to fixed vocabulary-specific heads, requiring manual relabeling and retraining whenever code sets change. We decoupled clinical document reading from target ontology indexing so one trained encoder can switch across arbitrary medical vocabularies in seconds without target training.

\noindent\textbf{Materials and Methods:} We developed Ontology-Compiled 2D Clinical State Fields (OCSF). A single-pass bidirectional diffusion-pretrained encoder and vocabulary-free decoder map each clinical note into a fixed-dimension two-dimensional state matrix with zero code-specific parameters. Any target ontology is compiled offline on CPU from public text descriptions into a scale-invariant Tikhonov deconvolution operator that suppresses sibling-code interference while preserving source decision thresholds. Models were trained solely on English MIMIC-IV ICD-9 discharge summaries and evaluated without target tuning across 58,706 held-out test notes and five unseen ontologies spanning English ICD-10, MS-DRG, Spanish, Greek, and Chinese (115--185,458 codes).

\noindent\textbf{Results:} On in-domain ICD-9 coding, OCSF reached 63.15\% Micro-F1, outperforming a matched closed-vocabulary classifier and eliminating late-epoch tail overfitting. Without target training, OCSF transferred directly to 26,085 English ICD-10 codes (29.23\% recall at gold cardinality), MS-DRG groups, and Spanish, Greek, and Chinese terminologies, compiling up to 185,458 codes in under 10 seconds on CPU with sub-millisecond GPU readout.

\noindent\textbf{Discussion:} Multi-channel partitioning and unit self-response normalization resolve concurrent comorbidity interference and preserve cross-lingual calibration.

\noindent\textbf{Conclusion:} OCSF turns clinical coding from static classification into instant, zero-training ontology compilation across healthcare standards and languages.

\vspace{0.35em}
\end{abstract}

\section{Background and Significance}
% =========================================================================

Healthcare institutions never operate under a permanently fixed medical vocabulary. National health authorities issue annual updates to diagnostic, procedural, and Diagnosis-Related Group (DRG) reimbursement schedules; health ministries worldwide are actively transitioning from legacy International Classification of Diseases (ICD-9 and ICD-10) editions toward ICD-11; and international research consortia must harmonize patient cohorts across languages and regional coding conventions~\cite{johnson-etal-2023-mimiciv,miranda-escalada-etal-2020-codiesp,dimitriadis-etal-2025-elcardiocc,zhang-etal-2022-cblue}. Within a single hospital stay, the same discharge summary is routinely mapped simultaneously to granular clinical codes for epidemiology and consolidated billing groups for hospital reimbursement. At the same time, strict patient-privacy governance prohibits exporting inpatient notes outside hospital firewalls for repeated external annotation. In operational deployment, an automated coding system must therefore switch target vocabularies in seconds using only publicly released codebooks, without collecting new labeled patient records or retraining neural parameters.

Existing clinical coding architectures cannot support rapid vocabulary switching because they entangle document comprehension with fixed label identifiers or incur prohibitive inference cost. Discriminative coders, spanning convolutional label attention~\cite{mullenbach-etal-2018-caml,vu-etal-2020-laat,li-yu-2020-multirescnn}, domain-pretrained Transformers~\cite{huang-etal-2022-plm,edin-etal-2023-automated,yuan-etal-2022-msmn}, and relational or traceable extensions~\cite{luo-etal-2024-corelation,ren-etal-2025-tracecoder}, allocate a dedicated parameter vector to every source code identifier; when a taxonomy changes, the classification head becomes incompatible, and multi-version retraining still requires labeled notes in the new taxonomy~\cite{liu-nguyen-2026-duallaat}. Generative large language models remove fixed heads by prompting over candidate descriptions or generating structured semantic identifiers~\cite{yang-etal-2023-gpsoap,boyle-etal-2023-llm,cheng-etal-2026-semicd}, yet clinical coding audits show substantial hallucination and limited recall across large codebooks unless paired with heavy retrieval pipelines~\cite{soroush-etal-2024-nejmai,klang-etal-2025-nejmai}. Bi-encoder retrieval and one-dimensional analytic weight projection~\cite{rios-kavuluru-2018-fewshot,heyden-etal-2026-cast} match a single document vector against public code embeddings, but inpatient discharge summaries average over thirteen concurrent diagnoses and procedures across multiple organ systems. Compressing an entire multi-system stay into one vector blends distinct findings together, while sibling codes with nearly identical wording trigger mutual false positives.

To enable instant vocabulary switching without losing multi-label precision, we separate reading the clinical narrative from querying the target codebook. During document reading, Ontology-Compiled 2D Clinical State Fields (OCSF) use a single-pass bidirectional diffusion-pretrained encoder~\cite{eslami-etal-2026-pplxembed,zhang-etal-2025-diffembed} and a vocabulary-free decoder to route clinical evidence from the full discharge summary into hundreds of semantic specialty channels. Storing concurrent conditions across distinct channels of a fixed-size two-dimensional state matrix prevents cardiopulmonary, renal, and infectious findings from cancelling one another in a single vector. Whenever a hospital switches to an unseen target ontology, OCSF compiles the public code descriptions offline on a standard CPU into a closed-form linear deconvolution operator. Rather than scoring each description in isolation, this compiled operator accounts for local similarity among sibling codes within each specialty channel to suppress cross-talk and standardize score magnitudes, reducing online inference on any compiled vocabulary to a sub-millisecond matrix multiplication.

This study introduces four research contributions for ontology-agnostic clinical coding:
\begin{enumerate}
  \item \textbf{Zero-Training Ontology Switching via 2D Clinical State Fields:} Clinical coding is formulated as predicting a vocabulary-free 2D state matrix, letting one English ICD-9 model switch immediately across revisions, billing schedules, and four languages without target supervision.
  \item \textbf{Scale-Invariant Deconvolution Compilation:} A closed-form Woodbury compiler with unit self-response normalization resolves sibling-code interference on CPU in seconds while preserving calibration across a $1{,}600\times$ vocabulary-size span.
  \item \textbf{Algebraic Compilation of Public Healthcare Rules:} Published authority rules (MS-DRG severity tiers, surgical/medical partitions, and procedural equivalence groups) compile algebraically into the readout operator with zero target parameters.
  \item \textbf{Mechanistic Characterization of Clinical Diffusion Pretraining:} Controlled comparisons against matched causal encoders and output-space flow models show that diffusion language models advance sparse clinical coding via full-document bidirectional representation pretraining rather than iterative output smoothing.
\end{enumerate}

% =========================================================================
\section{Materials and Methods}
% =========================================================================

\subsection{Problem Formulation and Cross-Ontology Evaluation Benchmarks}

Let source ontology $A$ contain $|A|$ clinical codes with public text descriptions $\{d_i^A\}_{i=1}^{|A|}$ and labeled pairs $\mathcal{D}_A = \{(x_n, y_n)\}_{n=1}^{N_A}$, where $x_n$ is a clinical note and $y_n \subseteq \{1,\dots,|A|\}$ is its ground-truth code set. At deployment, the system receives an unseen target ontology $B = \{(c_j, d_j^B)\}_{j=1}^{|B|}$ defined solely by public descriptions and optional structural rules, with zero labeled target notes ($\mathcal{D}_B^{\text{train}} = \varnothing$), and outputs calibrated scores $s_B(x) \in \mathbb{R}^{|B|}$ and predictions $\hat{y}_B(x)$ using source-calibrated parameters alone.

\begin{table}[htbp]
\centering
\scriptsize
\caption{\textbf{Overview of the source training ontology and five unseen target evaluation benchmarks.} OCSF is trained exclusively on English MIMIC-IV ICD-9 clinical notes ($154{,}542$ training notes across $72{,}216$ patients; $21{,}980$ validation notes across $10{,}068$ patients). Target ontologies span four languages, three orders of magnitude in vocabulary size ($115$ to $185{,}458$ codes), and both multi-label clinical coding and single-label reimbursement or chapter classification, with zero target training or threshold tuning.}
\label{tab:benchmarks}
\resizebox{\linewidth}{!}{%
\begin{tabular}{llllrrl}
\toprule
\textbf{Benchmark} & \textbf{Role} & \textbf{Language} & \textbf{Task Structure} & \textbf{$|B|$ (Codes)} & \textbf{Dev / Held-Out Test} & \textbf{Cohort Isolation \& Evaluation Rule} \\
\midrule
MIMIC-IV ICD-9 & Source ($A$) & English & Multi-label diag.\ + proc. & 11,324 (6,150 freq.) & 21,980 / 32,804 & Patient-disjoint splits; calibrates $\tau_A$ and $\bar{U}_A$ \\
\midrule
MIMIC-IV ICD-10 & Unseen ($B_1$) & English & Multi-label CM + 7-axis PCS & 26,085 (18,143 rare) & 9,077 / 13,783 & Excludes all 82,284 source train/val patients \\
MIMIC-IV MS-DRG & Unseen ($B_2$) & English & Single-label billing group & 1,581 (CMS v25--v37) & 7,875 / 11,869 & Excludes all 82,284 source train/val patients \\
CodiEsp & Unseen ($B_3$) & Spanish & Multi-label ICD-10-ES & 185,458 (11,381 OR) & 250 / 250 & Independent Spanish clinical case corpus \\
ElCardioCC & Unseen ($B_4$) & Greek & Single-label ICD-10 chapter & 115 (D50--D64) & 2,500 / ---$^{\dagger}$ & Independent Greek cardiology discharge notes \\
CHIP-CDN & Unseen ($B_5$) & Chinese & Multi-label ICD-10-CN terms & 37,645 & 2,000 / ---$^{\dagger}$ & OOV gold labels retained in recall denominator \\
\bottomrule
\multicolumn{7}{l}{\tiny $^{\dagger}$Publicly released with a single labeled evaluation cohort; evaluated strictly zero-shot without parameter updates or threshold calibration.} \\
\end{tabular}%
}
\end{table}

Table~\ref{tab:benchmarks} summarizes the source corpus and five unseen target benchmarks~\cite{johnson-etal-2023-mimiciv,edin-etal-2023-automated,miranda-escalada-etal-2020-codiesp,dimitriadis-etal-2025-elcardiocc,zhang-etal-2022-cblue}. Source training uses English MIMIC-IV v2.2 ICD-9 discharge summaries ($154{,}542$ training notes across $72{,}216$ patients; mean $1{,}782$ tokens, $13.43$ codes/note). All $82{,}284$ patients in ICD-9 training or validation are excluded from MIMIC-IV ICD-10 and MS-DRG cohorts to prevent patient overlap. Model choices, shrinkage $\lambda=0.15$, source threshold $\tau_A$, and mean source state $\bar{U}_A$ are fixed exclusively on ICD-9 validation. Target development splits serve only for zero-training evaluation, and the four held-out test splits ($58{,}706$ notes) were evaluated once in a single locked pass. On Chinese CHIP-CDN, gold codes absent from the target catalog remain in the recall denominator so out-of-vocabulary terms are penalized rather than discarded.

\subsection{Two-Dimensional Clinical State Representation and Deconvolution Compiler}

To decouple clinical note representation from target vocabulary size, let $f_{\text{cb}}: \mathcal{T} \to \mathbb{S}^{K-1}$ denote a frozen multilingual text encoder mapping any code description $d$ to a unit vector $z = f_{\text{cb}}(d) \in \mathbb{R}^K$ ($\|z\|_2 = 1$). Offline on source codebook embeddings $Z_A \in \mathbb{R}^{|A| \times K}$, we fit $R$ unit semantic channel anchors $A_c = [a_1,\dots,a_R]^\top \in \mathbb{R}^{R \times K}$ via spherical $k$-means, partitioning $A$ into disjoint specialty subsets $A_r = \{i \in A : \arg\max_{r'} a_{r'}^\top z_i = r\}$. Any multi-label code set $y \subseteq A$ then maps deterministically to a permutation-free two-dimensional ground-truth clinical state matrix $U_0(y) \in \mathbb{R}^{R \times K}$ via within-channel linear superposition:
\begin{equation}
  U_0(y)_r = \sum_{i \in y \cap A_r} z_i \in \mathbb{R}^K, \qquad r = 1,\dots,R.
  \label{eq:forward_state}
\end{equation}
At $R=512$ channels, the $11{,}324$ ICD-9 codes average $22.1$ codes per channel and the $13.43$ active codes per inpatient stay disperse across distinct rows, reducing expected within-channel active collisions below $1.03$ codes per active channel.

When switching to an unseen target ontology $B$, each description embedding $z_j = f_{\text{cb}}(d_j^B)$ routes deterministically to its nearest frozen source anchor $r(j) = \arg\max_r a_r^\top z_j$, forming channel submatrices $Z_{B,r} \in \mathbb{R}^{n_r \times K}$ ($\sum_{r=1}^R n_r = |B|$). Scoring $U_r$ via raw cosine products $Z_{B,r} U_r$ suffers severe false-positive cross-talk whenever sibling descriptions share high pairwise cosine similarity ($z_i^\top z_j \in [0.75, 0.96]$). Conversely, standard ridge inversion $(Z_{B,r}^\top Z_{B,r} + \lambda I_K)^{-1}$ fails across heterogeneous ontologies because $\operatorname{Tr}(Z_{B,r}^\top Z_{B,r}) = n_r$ scales linearly with channel occupancy $n_r$, shrinking logit magnitudes in dense ontologies such as 185k-code Spanish CodiEsp and invalidating source decision thresholds.

We resolve both failure modes via Scale-Invariant Self-Normalized Tikhonov Deconvolution. First, normalizing the within-channel Gram matrix by $n_r$,
\begin{equation}
  \bar{\Sigma}_{B,r} = \frac{1}{n_r} Z_{B,r}^\top Z_{B,r}, \qquad \operatorname{Tr}(\bar{\Sigma}_{B,r}) = 1,
  \label{eq:trace1_cov}
\end{equation}
makes $\lambda > 0$ a dimensionless shrinkage parameter with bounded condition number $\kappa(\bar{\Sigma}_{B,r} + \lambda I_K) \le (1+\lambda)/\lambda$ across arbitrary vocabulary sizes $|B|$. Second, defining raw deconvolved rows $\tilde{W}_{B,r} = Z_{B,r}(\bar{\Sigma}_{B,r} + \lambda I_K)^{-1}$ and self-response gains $g_{B,r,i} = (\tilde{W}_{B,r})_i^\top (Z_{B,r})_i > 0$, we normalize each row to exact unit self-response:
\begin{equation}
  W_{B,r} = \operatorname{diag}(g_{B,r})^{-1} Z_{B,r}\left(\bar{\Sigma}_{B,r} + \lambda I_K\right)^{-1} \in \mathbb{R}^{n_r \times K}.
  \label{eq:selfnorm_deconv}
\end{equation}
Under Eq.~\eqref{eq:selfnorm_deconv}, an isolated active concept $U_r = z_i$ yields an exact score of $w_i^\top z_i = 1.0$ regardless of ontology density $|B|$, while correlated inactive siblings $j \neq i$ are suppressed toward zero. Whenever $n_r < K$, we evaluate $\tilde{W}_{B,r}$ identically via the Woodbury dual $\tilde{W}_{B,r} = (\frac{1}{n_r} Z_{B,r} Z_{B,r}^\top + \lambda I_{n_r})^{-1} Z_{B,r}$, reducing compilation complexity from $\mathcal{O}(K^3)$ to $\mathcal{O}(n_r^3)$ (up to $1{,}000\times$ faster on CPU, maximum 32-bit floating-point deviation $2.38 \times 10^{-7}$). Algorithm~1 summarizes offline ontology compilation and online clinical state readout.

\begin{algorithm}[htbp]
\centering
\small
\begin{tabular}{p{0.96\linewidth}}
\toprule
\textbf{Algorithm 1: Offline Ontology Compilation and Online Clinical State Readout} \\
\midrule
\textbf{Input:} Frozen source anchors $A_c \in \mathbb{R}^{R \times K}$; public target codebook $B = \{(c_j, d_j^B)\}_{j=1}^{|B|}$; shrinkage $\lambda$; clinical note $x$. \\
\textbf{Output:} Compiled readout operator $W_B = \{W_{B,r}\}_{r=1}^R$ and zero-training target predictions $\hat{y}_B(x)$. \\[0.3em]
\textbf{Offline Ontology Compilation (CPU, executed once per codebook $B$):} \\
\quad Encode public descriptions $z_j \leftarrow f_{\text{cb}}(d_j^B) \in \mathbb{S}^{K-1}$ and assign channels $r(j) \leftarrow \arg\max_{r \in \{1..R\}} a_r^\top z_j$. \\
\quad \textbf{for} $r = 1,\dots,R$ \textbf{with} $n_r = |\{j : r(j)=r\}| > 0$ \textbf{do} \\
\quad\quad Form $Z_{B,r} \in \mathbb{R}^{n_r \times K}$. \textbf{If} $n_r < K$, solve $\tilde{W}_{B,r} \leftarrow \big(\frac{1}{n_r} Z_{B,r} Z_{B,r}^\top + \lambda I_{n_r}\big)^{-1} Z_{B,r}$; \\
\quad\quad \textbf{else} solve $\tilde{W}_{B,r} \leftarrow Z_{B,r}\big(\frac{1}{n_r} Z_{B,r}^\top Z_{B,r} + \lambda I_K\big)^{-1}$. \\
\quad\quad Compute gains $g_{B,r,i} \leftarrow (\tilde{W}_{B,r})_i^\top (Z_{B,r})_i$ and normalize $W_{B,r} \leftarrow \operatorname{diag}(g_{B,r})^{-1}\tilde{W}_{B,r}$. \\
\quad \textbf{end for} \\[0.3em]
\textbf{Online Clinical Readout (GPU, single forward pass per note $x$):} \\
\quad Predict vocabulary-free 2D state matrix $\hat{U}(x) \in \mathbb{R}^{R \times K}$ via bidirectional encoder and channel decoder. \\
\quad Evaluate deconvolved scores $s_j(x) \leftarrow (W_{B,r(j)})_j^\top \hat{U}(x)_{r(j)}$ for all $j=1,\dots,|B|$ in parallel. \\
\quad Emit $\hat{y}_B(x) \leftarrow \{c_j : s_j(x) \ge \tau_A\}$ for multi-label coding or $\arg\max_j s_j(x)$ for single-label grouping. \\
\bottomrule
\end{tabular}
\end{algorithm}

\subsection{Bidirectional Diffusion Encoding and Vocabulary-Free State Decoding}

Discharge summaries scatter mutually disambiguating clinical evidence across thousands of tokens: an initial differential diagnosis in the admission history is frequently confirmed, refined, or ruled out by microbiology cultures and discharge medications pages later. Causal autoregressive encoders impose a lower-triangular mask $M_{\ell m} = -\infty \cdot \mathbb{I}[m > \ell]$ that prevents early clinical mentions from attending to downstream conclusions. OCSF instead encodes full discharge summaries up to $L=3{,}072$ tokens in a single unchunked pass using open-weight bidirectional diffusion-pretrained Transformer encoders at $0.6\text{B}$ ($28$ layers, $K=1{,}024$) and $4\text{B}$ ($36$ layers, $K=2{,}560$) scales, paired against architecture-matched causal encoders with identical layer counts, hidden widths, and tokenizers.\footnote{Open-weight bidirectional diffusion encoders: \url{https://huggingface.co/perplexity-ai/pplx-embed-context-v1-0.6B} and \url{https://huggingface.co/perplexity-ai/pplx-embed-context-v1-4B} (MIT License)~\cite{eslami-etal-2026-pplxembed}. Architecture-matched causal autoregressive baselines: \url{https://huggingface.co/Qwen/Qwen3-Embedding-0.6B} and \url{https://huggingface.co/Qwen/Qwen3-Embedding-4B} (Apache-2.0 License)~\cite{zhang-etal-2025-qwen3embed}.} We extract continuous pre-quantization contextual token states $H = \operatorname{L2Norm}(\tanh(h)) \in \mathbb{R}^{L \times K}$ prior to int8 quantization traps, freeze token embeddings and the bottom 50\% of Transformer blocks to preserve the pretrained multilingual manifold across Spanish, Greek, and Chinese, and adapt only the upper 50\% of blocks on English ICD-9 via Low-Rank Adaptation (LoRA; $r=32, \alpha=64$)~\cite{hu-etal-2022-lora}.

\begin{figure}[htbp]
  \centering
  \includegraphics[width=\textwidth]{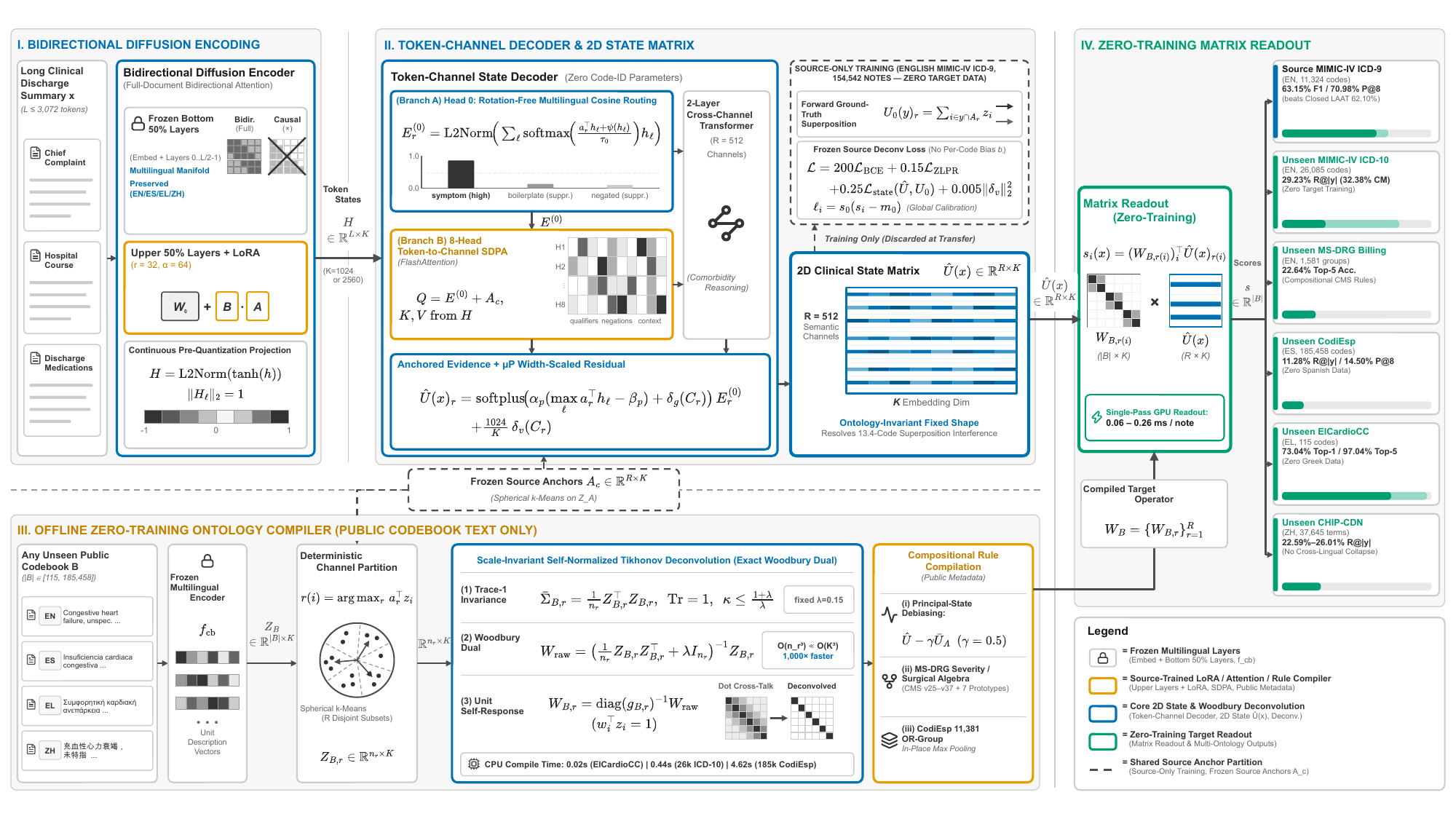}
  \caption{\textbf{Architecture of Ontology-Compiled 2D Clinical State Fields (OCSF).} \textbf{I:} A half-frozen bidirectional diffusion encoder maps an unchunked clinical narrative ($L \le 3{,}072$) into unit-norm contextual token states $H \in \mathbb{R}^{L \times K}$. \textbf{II:} A vocabulary-free token-channel decoder combines rotation-free cosine routing, multi-head token-to-channel attention, and cross-channel reasoning to emit an ontology-invariant 2D clinical state matrix $\hat{U}(x) \in \mathbb{R}^{R \times K}$. \textbf{III:} Offline on CPU, any unseen public codebook $B$ is compiled in $0.02\text{--}9.68\text{ s}$ via Scale-Invariant Self-Normalized Tikhonov Deconvolution (exact Woodbury dual) and optional compositional rule compilation. \textbf{IV:} Online inference reduces to a single sub-millisecond GPU matrix readout across six clinical ontologies.}
  \label{fig:overview}
\end{figure}

As illustrated in Figure~\ref{fig:overview}, the vocabulary-free state decoder maps $H \in \mathbb{R}^{L \times K}$ and frozen source anchors $A_c \in \mathbb{R}^{R \times K}$ to $\hat{U}(x) \in \mathbb{R}^{R \times K}$ with zero per-code parameters. First, rotation-free cosine routing pools multilingual token states directly in the frozen anchor basis,
\begin{equation}
  E^{(0)}_r = \operatorname{L2Norm}\!\left(\sum_{\ell=1}^L \operatorname{softmax}_\ell\!\left(\frac{a_r^\top h_\ell + \psi(h_\ell)}{\tau_0}\right) h_\ell\right),
  \label{eq:head0}
\end{equation}
where scalar salience $\psi: \mathbb{R}^K \to \mathbb{R}$ down-weights boilerplate phrasing without rotating non-English embeddings out of $f_{\text{cb}}$'s coordinate space. Second, 8-head scaled dot-product attention ($Q = E^{(0)} + A_c$, $K,V$ from $H$) followed by a 2-layer cross-channel Transformer ($R=512$) models contextual negation and inter-specialty comorbidities, emitting $\hat{U}(x)_r = \operatorname{softplus}(\alpha_p \max_\ell a_r^\top h_\ell - \beta_p + \delta_g(C_r)) E^{(0)}_r + \frac{1024}{K}\delta_v(C_r)$ with maximal-update ($\mu\text{P}$) width scaling $1024/K$~\cite{yang-etal-2021-mup} to prevent high-dimensional residual drift at $K=2{,}560$. Training on English ICD-9 minimizes a composite objective through the fixed source deconvolution operator $W_A$:
\begin{equation}
  \mathcal{L} = 200\,\mathcal{L}_{\text{BCE}}(\ell(x), y) + 0.15\,\mathcal{L}_{\text{ZLPR}}(\ell(x), y) + 0.25\,\mathcal{L}_{\text{state}}(\hat{U}(x), U_0(y)) + 0.005\,\|\delta_v\|_2^2,
  \label{eq:joint_loss}
\end{equation}
combining binary cross-entropy, zero-bounded log-sum-exp pairwise ranking ($\mathcal{L}_{\text{ZLPR}}$)~\cite{su-etal-2022-zlpr}, state alignment ($\mathcal{L}_{\text{state}}$), and two global calibration scalars $\ell_i(x) = s_0(w_{A,i}^\top \hat{U}(x)_{r(i)} - m_0)$ without per-code biases.

\subsection{Compositional Rule Compilation and Statistical Inference}

Public healthcare ontologies frequently publish explicit structural rules alongside code descriptions. Because OCSF's readout is linear in $\hat{U}(x)$, these rules compile algebraically into the target operator without learned target parameters:
\begin{enumerate}
  \item \textbf{Principal-Diagnosis State Debiasing:} Discharge summaries trained under exhaustive ICD-9 coding activate chronic background comorbidities alongside the acute admission driver. For single-label principal grouping (MS-DRG and Greek ElCardioCC), subtracting the mean source validation state $\tilde{U}(x) = \hat{U}(x) - \gamma \bar{U}_A$ ($\gamma=0.5$) suppresses chronic background prevalence prior to linear readout.
  \item \textbf{MS-DRG Severity and Surgical Rule Compilation:} Official Centers for Medicare \& Medicaid Services (CMS) MS-DRG v25--v37 tables decompose groups into base clinical conditions, three complication/comorbidity tiers (major, minor, or absent), and surgical-versus-medical partitions. Compiling seven public clinical concept prototypes via Eq.~\eqref{eq:selfnorm_deconv} adds zero-parameter structural bonuses ($w_{\text{sev}} = w_{\text{surg}} = 0.06$) to each group's base score.
  \item \textbf{Procedural Equivalence Pooling:} CodiEsp defines $11{,}381$ 4-character procedural equivalence prefixes over $7\text{-character}$ ICD-10-ES leaf codes; compiled max-pooling assigns each parent prefix the maximum deconvolved score among its constituent leaves.
\end{enumerate}

Multi-label benchmarks are evaluated primarily via recall at gold label cardinality ($\text{R@}|y|$, calibration-free) and zero-label thresholded $\text{Micro-F1}$ at source threshold $\tau_A$, supplemented by $\text{Precision@}k$ and mean average precision ($\text{mAP}$) in Appendix~\ref{sec:supp_dev_test}. Single-label benchmarks are evaluated via Top-1 and Top-5 accuracy. For every paired model comparison on identical clinical notes, we compute 2,000-resample paired bootstrap 95\% confidence intervals, two-sided $p$-values, and standardized effect sizes: paired Cohen's $d_z = \bar{\Delta}/s_\Delta$ (mean per-note gain divided by the sample standard deviation of paired differences) for continuous per-note metrics and Cohen's $h = 2\arcsin\sqrt{p_2} - 2\arcsin\sqrt{p_1}$ for single-label accuracy proportions.

% =========================================================================
\section{Results}
% =========================================================================

\subsection{Vocabulary-Free Decoding Prevents Tail Overfitting}

Replacing $17.93\text{M}$ per-code classifier weights with a fixed semantic deconvolution operator does not degrade in-domain coding performance; instead, OCSF surpasses closed-vocabulary baselines on the locked $32{,}804$-note MIMIC-IV ICD-9 test split (Table~\ref{tab:main_results}). On the standard Edin $6{,}150$-code benchmark~\cite{edin-etal-2023-automated}, half-frozen OCSF-0.6B outperforms the architecture-matched closed-vocabulary LAAT classifier trained under identical optimization settings ($\Delta = +0.27\text{ percentage points [pp]}$, $p < 0.0001$; Table~\ref{tab:stats_mechanisms}, Panel A), while full-LoRA OCSF-0.6B reaches $63.15\%$ Micro-F1 ($\Delta = +1.17\text{ pp}$, $95\%\text{ CI } [+1.05, +1.28]$, $d_z = +0.11$, $p < 0.0001$), exceeding published domain-pretrained PLM-ICD ($62.60\%$)~\cite{huang-etal-2022-plm,edin-etal-2023-automated}.

\begin{table}[htbp]
\centering
\scriptsize
\caption{\textbf{In-domain coding and zero-training cross-ontology transfer across six clinical ontologies.} ICD-9 ($N=32{,}804$; Edin 6,150 frequent codes), patient-disjoint ICD-10 ($N=13{,}783$; 26,085 codes split into 16,146 CM diagnoses and 9,939 PCS procedures), patient-disjoint MS-DRG ($N=11{,}869$), and Spanish CodiEsp ($N=250$) are evaluated on locked held-out test splits; Greek ElCardioCC ($N=2{,}500$) and Chinese CHIP-CDN ($N=2{,}000$) are evaluated on their full public cohorts. All thresholds $\tau_A$, channel anchors, and $\lambda=0.15$ are fixed solely on source ICD-9 validation. Values are percentages (full precision/ranking metrics and development-to-test consistency in Supplementary Tables S1--S2).}
\label{tab:main_results}
\resizebox{\linewidth}{!}{%
\begin{tabular}{lrrrrrrrrr}
\toprule
 & \textbf{ICD-9} & \multicolumn{4}{c}{\textbf{ICD-10 (Zero-Training, 26,085)}} & \textbf{MS-DRG} & \textbf{CodiEsp} & \textbf{ElCardioCC} & \textbf{CHIP-CDN} \\
\cmidrule(lr){3-6}
\textbf{Method} & F1 & All R@$|y|$ & All F1 & CM R@$|y|$ & PCS R@$|y|$ & Top-1/5 & ES R@$|y|$ & EL Top-1/5 & ZH R@$|y|$ \\
\midrule
\multicolumn{10}{l}{\textit{Closed-Vocabulary Supervised Baselines (Bound to ICD-9 Head; Cannot Score Unseen Ontologies)}} \\
PLM-ICD\cite{edin-etal-2023-automated} (published) & 62.60 & --- & --- & --- & --- & --- & --- & --- & --- \\
Closed-Vocabulary LAAT (Diffusion-0.6B, matched) & 62.10 & --- & --- & --- & --- & --- & --- & --- & --- \\
\midrule
\multicolumn{10}{l}{\textit{Zero-Shot Bi-Encoder \& 1D Cross-Ontology Projection Baselines}} \\
Zero-Shot Bi-Encoder (Diffusion-0.6B) & 5.02 & 1.66 & 1.17 & --- & --- & 3.45/13.40 & 3.42 & 33.12/78.72 & 25.87 \\
Zero-Shot Bi-Encoder (Diffusion-4B) & 6.46 & 3.33 & 2.17 & --- & --- & 4.23/17.87 & 3.63 & 50.52/90.48 & 25.23 \\
Zero-Shot Bi-Encoder (Causal-4B) & 7.31 & 5.75 & 3.00 & --- & --- & 9.71/26.73 & 3.12 & 45.76/86.64 & 30.17 \\
Linear Weight Projection (1D, Diffusion-0.6B)$^{\ddagger}$ & --- & 0.11 & 0.35 & --- & --- & 1.49/2.25 & 0.05 & 0.16/39.64 & 0.06 \\
\midrule
\multicolumn{10}{l}{\textit{Ontology-Compiled 2D Clinical State Fields --- 0.6B Scale ($K=1{,}024$)}} \\
OCSF (Diffusion-0.6B, $R=1$ 1D State) & 33.69 & 17.60 & 10.72 & 19.02 & 3.64 & 6.81/23.73 & 7.94 & 20.28/76.40 & 22.70 \\
OCSF (Diffusion-0.6B, $R=128$) & 61.15 & 24.86 & 20.92 & 26.84 & 7.10 & 6.80/21.47 & 10.50 & 24.52/72.00 & 20.87 \\
OCSF (Diffusion-0.6B, $R=512$, Cosine Readout) & 27.24 & 18.50 & 9.19 & 20.28 & 6.32 & 5.16/15.42 & 7.90 & 36.44/76.96 & 21.15 \\
OCSF (Diffusion-0.6B, $R=512$, Unnormalized Deconv.) & 26.35 & 26.19 & 12.42 & 28.05 & 6.87 & 5.09/17.90 & 9.72 & 22.88/73.72 & 21.62 \\
OCSF (Causal-0.6B, $R=512$, Half-Frozen) & 57.75 & 23.14 & 20.72 & 25.75 & 5.98 & 7.09/20.27 & 6.45 & 30.24/69.84 & 18.75 \\
\textbf{OCSF (Diffusion-0.6B, $R=512$, Half-Frozen)} & \textbf{62.25} & \textbf{27.97} & \textbf{22.82} & \textbf{30.55} & \textbf{7.39} & \textbf{6.37}/\textbf{21.48} & \textbf{10.47} & \textbf{42.20}/\textbf{79.40} & \textbf{22.59} \\
\textbf{OCSF (Diffusion-0.6B, $R=512$, Full-LoRA)} & \textbf{63.15} & \textbf{28.66} & \textbf{24.10} & \textbf{31.29} & \textbf{7.55} & \textbf{6.95}/\textbf{21.94} & \textbf{9.59} & \textbf{45.44}/\textbf{80.12} & \textbf{21.15} \\
\midrule
\multicolumn{10}{l}{\textit{Ontology-Compiled 2D Clinical State Fields --- 4B Scale ($K=2{,}560$)}} \\
OCSF (Causal-4B, $R=512$, Half-Frozen) & 59.12 & 26.55 & 23.72 & 29.91 & 9.50 & 9.91/24.92 & 10.74 & 60.80/95.72 & 26.01 \\
OCSF (Diffusion-4B, $R=512$, Unnormalized Deconv.) & 24.98 & 27.93 & 9.76 & 30.69 & 6.64 & 5.22/17.74 & 10.93 & --- & --- \\
OCSF (Diffusion-4B, $R=512$, Half-Frozen) & 60.09 & 29.05 & 25.98 & 32.25 & 8.65 & 7.42/22.28 & 10.89 & 73.04/97.04 & 18.68 \\
\textbf{OCSF (Diffusion-4B, $R=512$, Smooth State)} & \textbf{60.21} & \textbf{29.23} & \textbf{26.12} & \textbf{32.38} & \textbf{8.34} & \textbf{7.41}/\textbf{22.64} & \textbf{11.28} & \textbf{70.44}/\textbf{97.04} & \textbf{19.92} \\
\bottomrule
\multicolumn{10}{l}{\tiny $^{\ddagger}$Evaluated on development splits (see Supplementary Table S1) to illustrate cross-lingual rank collapse of 1D linear readout projections ($0.06\%$ on Chinese).} \\
\end{tabular}%
}
\end{table}

Inspection of epoch-by-epoch validation trajectories reveals why removing per-code parameters improves generalization (Appendix~\ref{sec:supp_dev_test}). Closed-vocabulary LAAT peaks at Epoch 3 ($61.89\%$ validation Micro-F1) and steadily deteriorates to $61.37\%$ by Epoch 6 as unconstrained per-code weight vectors memorize scarce training examples in the $11{,}324$-code long tail. By contrast, OCSF improves monotonically through Epoch 6 ($62.11\%$ validation, $62.25\%$ test), because scoring every code through frozen textual coordinates $Z_A$ acts as a structural semantic regularizer that anchors rare codes to clinically related frequent diagnoses.

\subsection{Instant Zero-Training Switching Across Revisions, Billing Systems, and Languages}

Ordered by increasing ontological distance, from same-language taxonomic revision (ICD-10) to reimbursement synthesis (MS-DRG) and cross-lingual terminologies (Spanish, Greek, and Chinese), a single English ICD-9 model switches across all five unseen ontologies with zero target training (Table~\ref{tab:main_results}). Offline Woodbury compilation completes in $0.016\text{ s}$ for Greek ElCardioCC, $0.436\text{--}0.800\text{ s}$ for 26k-code ICD-10, and $4.62\text{--}9.68\text{ s}$ for 185k-code Spanish CodiEsp on CPU, while online GPU matrix readout requires only $0.05\text{--}0.26\text{ ms}$ per note (Table~\ref{tab:efficiency}). End-to-end single-pass inference across $2{,}048$-token discharge summaries runs at $53.0\text{ notes/s}$ in $2.86\text{ GB}$ VRAM at $0.6\text{B}$ and $15.6\text{ notes/s}$ in $12.67\text{ GB}$ VRAM at $4\text{B}$, fitting comfortably on a single hospital workstation GPU.

Across the $13{,}783$ patient-disjoint MIMIC-IV ICD-10 test notes, OCSF-4B lifts zero-training recall nearly ninefold over zero-shot bi-encoder retrieval ($29.23\%$ vs.\ $3.33\%$ $\text{R@}|y|$) and achieves $26.12\%$ zero-label thresholded Micro-F1 using the unmodified ICD-9 threshold $\tau_A$. Stratifying ICD-10 into its clinical sub-systems uncovers a sharp structural boundary between diagnostic and procedural transfer: OCSF-4B recovers $32.38\%$ $\text{R@}|y|$ ($29.54\%$ Micro-F1) across $16{,}146$ ICD-10-CM diagnostic codes, compared with $8.34\%$ across $9{,}939$ ICD-10-PCS surgical codes. Unlike diagnostic narratives whose clinical semantics carry directly across ICD revisions, 7-character multiaxial ICD-10-PCS codes specify exact anatomical approach, device, and qualifier axes that were never annotated in legacy 3-to-4-digit ICD-9 procedure records.

Cross-lingual transfer exposes a second structural phenomenon: one-dimensional cross-ontology projections suffer catastrophic rank collapse when crossing language families, whereas two-dimensional channel routing preserves the pretrained multilingual manifold. On Chinese CHIP-CDN, fitting a 1D linear readout projection on English ICD-9 collapses recall from $25.87\%$ (zero-shot) to $0.06\%$ (Table~\ref{tab:main_results}), because an unconstrained English linear map projects non-English embeddings into its nullspace. By routing contextual states via rotation-free cosine similarity in the frozen anchor basis and freezing the bottom 50\% of encoder layers, OCSF transfers zero-shot from English ICD-9 to $185{,}458$ Spanish codes ($11.28\%$ $\text{R@}|y|$), Greek cardiology chapters ($73.04\%$ Top-1 and $97.04\%$ Top-5 accuracy), and Chinese clinical terms ($22.59\%$--$26.01\%$ $\text{R@}|y|$). Unfreezing all encoder layers (Full-LoRA) further boosts English ICD-9 ($+0.90\text{ pp}$) and English ICD-10 ($+0.69\text{ pp}$) but degrades Spanish CodiEsp transfer from $10.47\%$ to $9.59\%$ ($d_z = -0.12$, $p=0.059$), confirming that bottom-layer freezing actively shields non-English representations from English-specific drift.

Algebraically compiling public healthcare rules yields consistent zero-parameter gains whenever target tasks depart from flat multi-label coding (Table~\ref{tab:stats_mechanisms}, Panel A). On $11{,}869$ patient-disjoint MS-DRG test admissions, compiling CMS severity/surgical rules with principal-diagnosis debiasing raises Top-1 accuracy by $+1.14\text{ pp}$ ($18\%$ relative gain over pure description deconvolution; $d_z = +0.06, h = +0.05, p < 0.0001$), improving accuracy across all $13$ annual CMS grouper versions from v25 through v37 (Appendix Table~\ref{tab:supp_drg_and_noise}). The same principal-state debiasing lifts Greek ElCardioCC Top-1 accuracy by $+5.32\text{ pp}$ ($d_z = +0.12, h = +0.11, p < 0.0001$), while compiled OR-group pooling improves Spanish CodiEsp recall ($d_z = +0.16, p = 0.007$).

\begin{table}[htbp]
\centering
\scriptsize
\caption{\textbf{Statistical inference with standardized effect sizes, multi-seed channel-partitioning stability, and output-flow mechanism analysis.} Panel A reports per-note paired differences ($\Delta$, percentage points), 95\% bootstrap confidence intervals, paired Cohen's $d_z = \bar{\Delta}/s_{\Delta}$ (with Cohen's $h$ for single-label accuracy), and two-sided $p$-values. Panel B reports 5-seed mean $\pm$ SD noisy-state recovery across partitioning rules ($R=512$). Panel C compares direct single-pass state readout ($N=0$) against $N$-step output-space Rectified Flow smoothing.}
\label{tab:stats_mechanisms}
\resizebox{\linewidth}{!}{%
\begin{tabular}{llrrrr}
\toprule
\textbf{Comparison} & \textbf{Cohort ($N$)} & \textbf{$\Delta$ (pp)} & \textbf{95\% CI} & \textbf{Cohen's $d_z$} & \textbf{$p$-value} \\
\midrule
\multicolumn{6}{l}{\textbf{Panel A: Paired Bootstrap Significance and Standardized Effect Sizes (2,000 Resamples)}} \\
OCSF (Diffusion-0.6B, Full-LoRA) vs.\ Closed LAAT & ICD-9 Freq Test (32,804) & +1.17 & [+1.05, +1.28] & +0.11 & $<0.0001$ \\
OCSF (Diffusion-0.6B, Half-Frozen) vs.\ Closed LAAT & ICD-9 Freq Test (32,804) & +0.27 & [+0.16, +0.39] & +0.03 & $<0.0001$ \\
Channel Expansion ($R=512$ vs.\ $R=1$, 0.6B) & ICD-9 Freq Test (32,804) & +22.02 & [+21.85, +22.18] & +1.44 & $<0.0001$ \\
Channel Expansion ($R=512$ vs.\ $R=1$, 0.6B) & ICD-10 All Test (13,783) & +10.37 & [+10.16, +10.58] & +0.81 & $<0.0001$ \\
Self-Norm.\ Deconv.\ vs.\ Cosine Readout (4B) & ICD-10 All Test (13,783) & +10.46 & [+10.26, +10.66] & +0.90 & $<0.0001$ \\
Self-Norm.\ vs.\ Unnormalized Deconv.\ (0.6B) & ICD-9 Freq Test (32,804) & +8.34 & [+8.20, +8.47] & +0.67 & $<0.0001$ \\
Self-Norm.\ vs.\ Unnormalized Deconv.\ (4B) & ICD-10 All Test (13,783) & +1.30 & [+1.16, +1.44] & +0.15 & $<0.0001$ \\
Structural Rule Compilation vs.\ Pure Deconv.\ (4B) & MS-DRG Test (11,869) & +1.14 & [+0.82, +1.48] & +0.06 ($h=+0.05$) & $<0.0001$ \\
Structural Rule Compilation vs.\ Pure Deconv.\ (0.6B) & ElCardioCC EL (2,500) & +5.32 & [+3.56, +7.00] & +0.12 ($h=+0.11$) & $<0.0001$ \\
Bidirectional Diffusion vs.\ Causal (0.6B) & ICD-10 All Test (13,783) & +4.83 & [+4.63, +5.02] & +0.41 & $<0.0001$ \\
Bidirectional Diffusion vs.\ Causal (0.6B) & CodiEsp ES Test (250) & +4.02 & [+2.81, +5.23] & +0.40 & $<0.0001$ \\
Bidirectional Diffusion vs.\ Causal (4B) & ICD-10 All Test (13,783) & +2.68 & [+2.47, +2.89] & +0.22 & $<0.0001$ \\
Bidirectional Diffusion vs.\ Causal (4B) & ElCardioCC EL (2,500) & +12.24 & [+10.04, +14.56] & +0.22 ($h=+0.26$) & $<0.0001$ \\
Scale 4B vs.\ 0.6B (Bidirectional Diffusion) & ICD-10 CM Test (13,782) & +1.83 & [+1.62, +2.04] & +0.14 & $<0.0001$ \\
\midrule
\multicolumn{6}{l}{\textbf{Panel B: Multi-Seed Channel-Partitioning Stability under Noisy State Recovery (Mean $\pm$ SD across 5 Seeds)}} \\
\textit{Partition Strategy} & \textit{Setting} & \textit{ICD-9 R@$|y|$ (\%)} & \textit{ICD-10 R@$|y|$ (\%)} & \textit{Max Load $n_r$} & \textit{Empty Ch.} \\
Spherical $k$-Means (Default) & 5 Seeds ($R=512$) & 73.78$\pm$0.71 & 48.44$\pm$0.80 & 654.2$\pm$106.5 & 17.8 \\
Hierarchical Ward Linkage & 5 Seeds ($R=512$) & 74.83$\pm$0.32 & 47.85$\pm$0.88 & 677.0$\pm$0.0 & 10.0 \\
Uniform Random Source Anchors & 5 Seeds ($R=512$) & 74.89$\pm$1.41 & 51.55$\pm$1.43 & 641.8$\pm$81.6 & 2.2 \\
\midrule
\multicolumn{6}{l}{\textbf{Panel C: Representation Pretraining vs.\ Output-Space Rectified Flow Steps ($N \in \{0,2,4,8\}$, Diffusion-0.6B Dev)}} \\
\textit{Inference Mode} & \textit{Euler Steps} & \textit{ICD-9 Val R@$|y|$} & \textit{ICD-10 Dev R@$|y|$} & \textit{CHIP-CDN R@$|y|$} & \textit{Regime} \\
Output-Space Rectified Flow & $N=0$ (Direct) & 23.64 & 6.70 & 23.98 & Single-pass readout \\
Output-Space Rectified Flow & $N=2$ & 21.15 & 6.77 & 16.07 & Iterative flow \\
Output-Space Rectified Flow & $N=4$ & 21.77 & 6.91 & 15.02 & Iterative flow \\
Output-Space Rectified Flow & $N=8$ & 22.08 & 7.04 & 14.77 & Iterative flow \\
\bottomrule
\end{tabular}%
}
\end{table}

\subsection{Channel Geometry and Self-Normalization Govern Capacity and Calibration}

Standardized effect sizes in Table~\ref{tab:stats_mechanisms} (Panel A) distinguish primary architectural drivers from minor large-sample differences: expanding state capacity from $R=1$ to $R=512$ channels ($d_z = +1.44$ on ICD-9; $d_z = +0.81$ on ICD-10) and replacing 2D cosine matching with self-normalized deconvolution ($d_z = +1.20$ on ICD-9; $d_z = +0.90$ on ICD-10) produce the largest effect sizes in the study.

\begin{figure}[htbp]
  \centering
  \includegraphics[width=\textwidth]{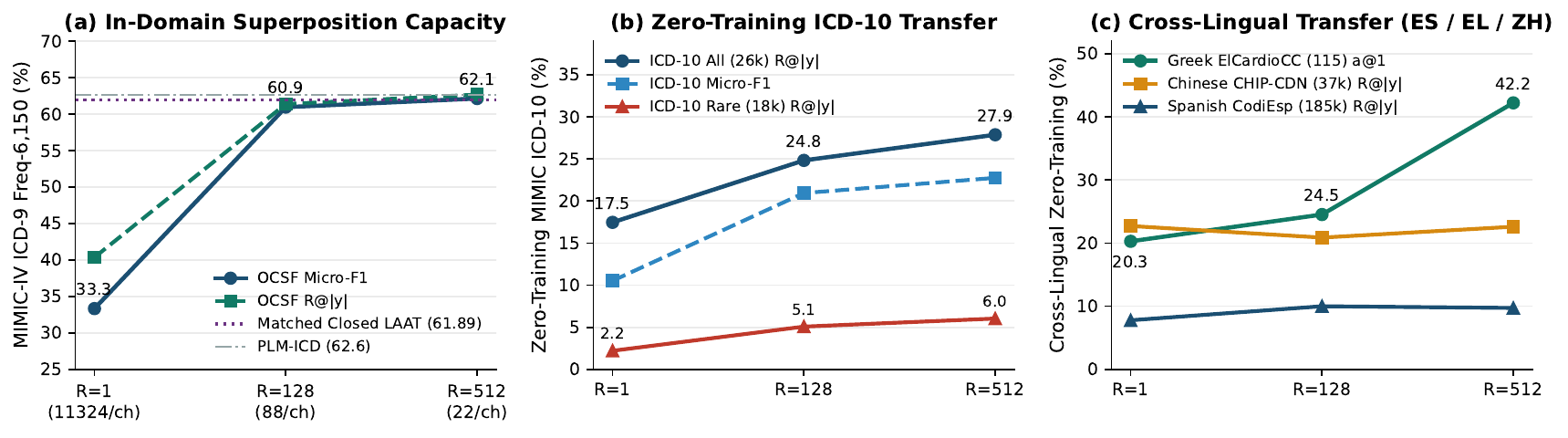}
  \caption{\textbf{Multi-label superposition scaling law across $R \in \{1, 128, 512\}$ semantic channels (Diffusion-0.6B).} \textbf{(a)} In-domain MIMIC-IV ICD-9 performance rises by $+28.6$ Micro-F1 points from 1D ($R=1$) to $R=512$, surpassing closed-vocabulary LAAT. \textbf{(b)} Zero-training ICD-10 recall rises by $+10.4$ points overall and nearly triples on $18{,}143$ rare codes. \textbf{(c)} Cross-lingual transfer improves monotonically across Greek ElCardioCC and Spanish CodiEsp without Chinese collapse.}
  \label{fig:scaling}
\end{figure}

Figure~\ref{fig:scaling} shows how two-dimensional channel expansion resolves comorbidity superposition. Compressing $13.4$ concurrent diagnoses into one $R=1$ vector caps ICD-9 test Micro-F1 at $33.69\%$ and unseen rare ICD-10 recall at $1.61\%$; partitioning into $R=512$ channels restores $62.25\%$ ICD-9 Micro-F1 and more than triples rare ICD-10 recall to $5.53\%$. Single-label MS-DRG exhibits the inverse trend, peaking at $R=1$ ($6.81\%\text{/}23.73\%$ Top-1/Top-5 vs.\ $6.37\%\text{/}21.48\%$ at $R=512$; $d_z = -0.01, p=0.13$). Because each admission carries only one DRG group, multi-label superposition is absent, confirming that $R > 1$ specifically resolves concurrent label interference rather than merely adding parameters. Across five random seeds (Table~\ref{tab:stats_mechanisms}, Panel B), spherical $k$-means achieves $73.78 \pm 0.71\%$ ICD-9 and $48.44 \pm 0.80\%$ ICD-10 noisy-state recovery, matching Ward linkage ($74.83 \pm 0.32\%$) with half the variance of random source anchors ($74.89 \pm 1.41\%$ and $51.55 \pm 1.43\%$).

\begin{figure}[htbp]
  \centering
  \includegraphics[width=\textwidth]{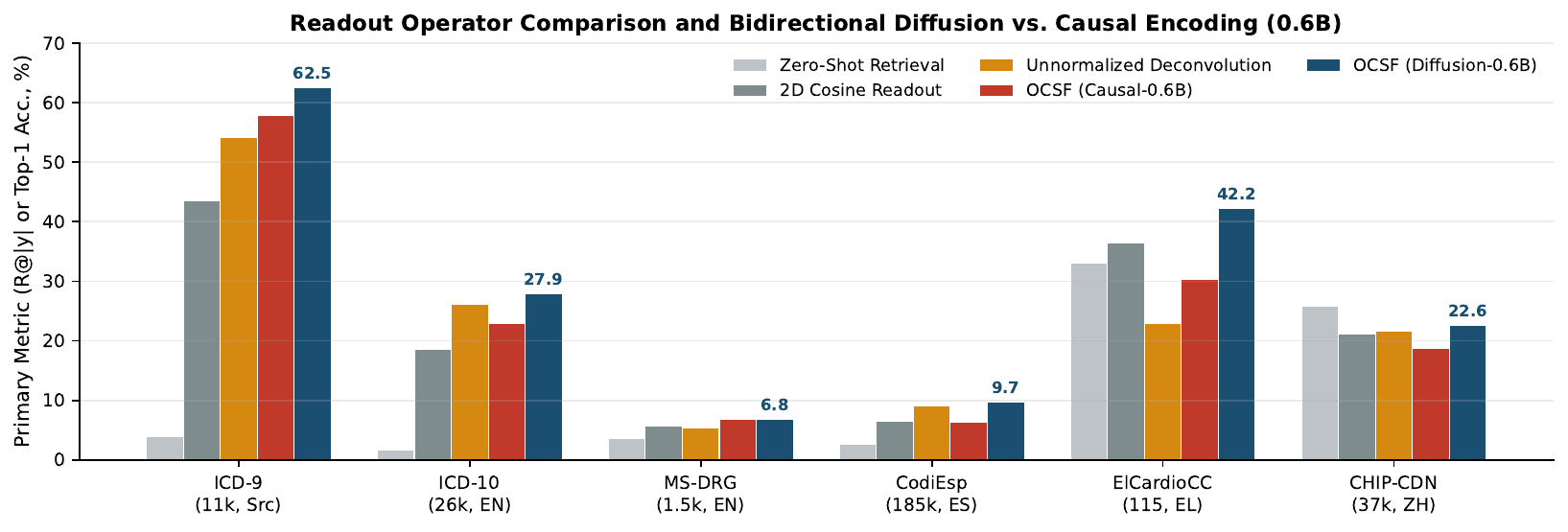}
  \caption{\textbf{Readout operator comparison and architecture-matched bidirectional diffusion vs.\ causal encoding (0.6B, $R=512$).} Within the exact same predicted state matrix $\hat{U}(x)$, replacing 2D cosine matching with Scale-Invariant Self-Normalized Deconvolution eliminates sibling cross-talk across dense vocabularies, while bidirectional diffusion encoding consistently outperforms causal encoding across five of six ontologies.}
  \label{fig:deconv_vs_causal}
\end{figure}

Unit self-response normalization ($w_i^\top z_i = 1$) proves essential for zero-label decision calibration across heterogeneous vocabulary sizes (Figure~\ref{fig:deconv_vs_causal} and Figure~\ref{fig:lambda_flow}a). Without row-wise self-normalization, raw Tikhonov deconvolution still improves ranking over cosine matching ($27.93\%$ vs.\ $18.77\%$ ICD-10 $\text{R@}|y|$ at $4\text{B}$) but collapses zero-label thresholded Micro-F1 from $26.12\%$ to $9.76\%$ on ICD-10 and from $8.33\%$ to $2.27\%$ on Spanish CodiEsp (Table~\ref{tab:main_results}) because channel density shifts raw logit scales away from $\tau_A$. Enforcing trace-normalized covariance (Eq.~\eqref{eq:trace1_cov}) and unit self-response (Eq.~\eqref{eq:selfnorm_deconv}) stabilizes a single dimensionless $\lambda=0.15$ at $\ge 97\%$ of per-target peak performance across a $1{,}600\times$ ontology-size span ($|B| \in [115, 185{,}458]$; Figure~\ref{fig:lambda_flow}a).

\begin{table}[htbp]
\centering
\scriptsize
\caption{\textbf{Computational efficiency of offline Woodbury dual compilation (CPU) and online inference (NVIDIA GH200 BF16 GPU, $R=512$).} Top: one-time offline codebook compilation latency on CPU and pure matrix readout latency per note across all six ontologies. Bottom: end-to-end single-pass latency (encoder + state decoder + 26,085-code ICD-10 readout) and peak GPU memory at typical ($L=2{,}048$) and maximum ($L=3{,}072$) discharge summary lengths.}
\label{tab:efficiency}
\resizebox{\linewidth}{!}{%
\begin{tabular}{lrrrrr}
\toprule
 & & \multicolumn{2}{c}{\textbf{Diffusion-0.6B ($K=1{,}024$)}} & \multicolumn{2}{c}{\textbf{Diffusion-4B ($K=2{,}560$)}} \\
\cmidrule(lr){3-4} \cmidrule(lr){5-6}
\textbf{Target Ontology} & \textbf{$|B|$} & \textbf{CPU Compile} & \textbf{GPU Readout} & \textbf{CPU Compile} & \textbf{GPU Readout} \\
\midrule
ElCardioCC (Greek) & 115 & 0.016\,s & 0.0596\,ms & 0.028\,s & 0.1316\,ms \\
MIMIC-IV MS-DRG (English) & 1,581 & 0.031\,s & 0.0583\,ms & 0.040\,s & 0.0481\,ms \\
MIMIC-IV ICD-9 (Source) & 11,324 & 0.145\,s & 0.1781\,ms & 0.268\,s & 0.1230\,ms \\
MIMIC-IV ICD-10 (English) & 26,085 & 0.436\,s & 0.1198\,ms & 0.800\,s & 0.1250\,ms \\
CHIP-CDN (Chinese) & 37,645 & 0.799\,s & 0.1208\,ms & 1.430\,s & 0.1218\,ms \\
CodiEsp (Spanish) & 185,458 & 4.616\,s & 0.2577\,ms & 9.681\,s & 0.1723\,ms \\
\midrule
\multicolumn{2}{l}{\textit{End-to-End Inference ($L=2{,}048$, batch 16, 26k ICD-10)}} & \multicolumn{2}{r}{18.89\,ms (53.0 notes/s, 2.86\,GB)} & \multicolumn{2}{r}{64.15\,ms (15.6 notes/s, 12.67\,GB)} \\
\multicolumn{2}{l}{\textit{End-to-End Inference ($L=3{,}072$, batch 16, 26k ICD-10)}} & \multicolumn{2}{r}{29.57\,ms (33.8 notes/s, 3.47\,GB)} & \multicolumn{2}{r}{102.27\,ms (9.8 notes/s, 14.33\,GB)} \\
\bottomrule
\end{tabular}%
}
\end{table}

\subsection{Bidirectional Diffusion Pretraining Versus Output-Space Smoothing}

Controlled comparisons against architecture-matched causal encoders isolate how diffusion language models benefit clinical coding (Figure~\ref{fig:deconv_vs_causal} and Table~\ref{tab:stats_mechanisms}, Panel A). At $0.6\text{B}$, bidirectional diffusion encoding outperforms causal encoding by $+4.48\text{ pp}$ Micro-F1 on ICD-9 ($d_z = +0.39$), $+4.83\text{ pp}$ $\text{R@}|y|$ on unseen ICD-10 ($d_z = +0.41$), $+4.02\text{ pp}$ on Spanish CodiEsp ($d_z = +0.40$), and $+11.96\text{ pp}$ Top-1 accuracy on Greek ElCardioCC ($h = +0.25$, all $p < 0.0001$). Scaling from $0.6\text{B}$ to $4\text{B}$ further improves zero-training ICD-10-CM recall ($32.38\%$ vs.\ $30.55\%$, $d_z = +0.14$) and lifts Greek cardiology accuracy to $73.04\%$ Top-1 / $97.04\%$ Top-5, maintaining bidirectional gains over causal $4\text{B}$ on ICD-10 ($+2.68\text{ pp}$, $d_z = +0.22$) and Greek ($+12.24\text{ pp}$, $h = +0.26$, $p < 0.0001$). Length stratification shows uniform bidirectional gains across short ($<1{,}200$ tokens: $+4.6\text{ pp}$), medium ($1{,}200\text{--}2{,}200$: $+5.0\text{ pp}$), and long ($>2{,}200$: $+4.9\text{ pp}$) notes (Appendix Table~\ref{tab:supp_length_and_ap}), confirming that bidirectional attention lets early admission findings and late hospital-course conclusions disambiguate each other globally.

\begin{figure}[htbp]
  \centering
  \includegraphics[width=\textwidth]{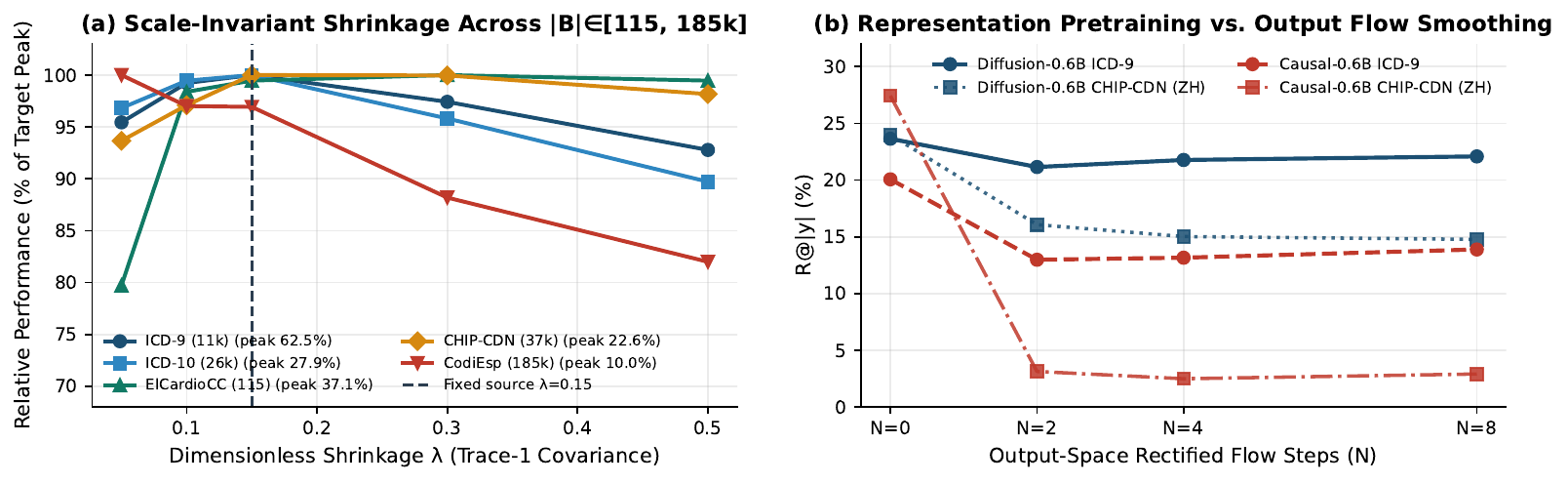}
  \caption{\textbf{Scale-invariant shrinkage transfer and mechanism of clinical diffusion models.} \textbf{(a)} Under trace-1 within-channel covariance, fixed source $\lambda=0.15$ achieves $\ge 97\%$ of per-target peak performance across a $1{,}600\times$ vocabulary range without target tuning. \textbf{(b)} Direct single-pass readout ($N=0$) from bidirectional diffusion representations outperforms $N$-step output-space Rectified Flow, which smooths sparse clinical code peaks toward conditional expectations.}
  \label{fig:lambda_flow}
\end{figure}

To test whether additional iterative denoising in the output state space could further refine predictions, we trained a conditional Rectified Flow velocity network~\cite{liu-etal-2023-rectifiedflow} over the 2D state matrix and swept $N \in \{0, 2, 4, 8\}$ Euler integration steps (Table~\ref{tab:stats_mechanisms}, Panel C and Figure~\ref{fig:lambda_flow}b). Stepping along deterministic ODE trajectories ($N > 0$) reduces recall on ICD-9 ($23.64\% \to 21.15\%$) and Chinese CHIP-CDN ($23.98\% \to 14.77\%$). Because ground-truth clinical states $U_0(y)$ are sparse discrete superpositions on unit-sphere facets, minimum-MSE velocity fields pull sharp code peaks toward conditional expectations over plausible sibling codes, attenuating rare diagnoses. Diffusion pretraining thus advances clinical coding through full-document bidirectional representation learning at the input rather than continuous smoothing at the output.

% =========================================================================
\section{Discussion}
% =========================================================================

Decoupling clinical narrative encoding from target ontology indexing removes a central deployment bottleneck in medical informatics. Annual CMS grouper revisions, national ICD transitions, and regional specialty registries ordinarily force hospitals into recurring cycles of chart abstraction, retraining, and recalibration. By encoding each stay into a fixed $512 \times K$ semantic state matrix and compiling public codebooks analytically on CPU in under ten seconds, OCSF enables instant terminology switching while keeping clinical text behind local firewalls.

Evaluating effect sizes alongside $p$-values clarifies component contributions across large cohorts. On splits exceeding $10{,}000$ admissions, differences under half a percentage point achieve $p < 0.0001$ while confirming practical parity ($d_z = 0.03$) rather than a large per-note shift. By contrast, multi-channel partitioning ($d_z = 0.81\text{--}1.44$), self-normalized deconvolution ($d_z = 0.84\text{--}1.20$), and bidirectional diffusion pretraining ($d_z = 0.22\text{--}0.41$) produce substantial per-note gains across English and non-English cohorts. Two architecture trade-offs also emerge. At $4\text{B}$, the causal encoder slightly exceeds the bidirectional encoder on MS-DRG ($9.91\%$ vs.\ $7.41\%$ Top-1) and Chinese CHIP-CDN ($26.01\%$ vs.\ $19.92\%$), reflecting stronger zero-shot lexical pretraining on English billing phrases and Chinese biomedical text in its base checkpoint ($9.71\%$ vs.\ $4.23\%$ on MS-DRG; $30.17\%$ vs.\ $25.23\%$ on CHIP-CDN before adaptation; Table~\ref{tab:main_results}). Scaling from $0.6\text{B}$ to $4\text{B}$ strengthens zero-training cross-ontology transfer across ICD-10, MS-DRG, Spanish, and Greek while slightly lowering in-domain ICD-9 F1 ($60.21\%$ vs.\ $62.25\%$), indicating milder source-taxonomy specialization in the wider frozen anchor space.

\subsection{Limitations}

Three boundaries define the scope of zero-training ontology compilation. First, although diagnostic transfer to unseen ICD-10-CM reaches $32.38\%$ recall, recall on 7-character ICD-10-PCS surgical procedures remains $8.34\%$ ($9.50\%$ with Causal-4B) because multiaxial surgical qualifiers (approach, device, qualifier) are absent from legacy ICD-9 supervision; bridging procedural axes will require compiling explicit multiaxial grammars from operative reports. Second, description-only compilation requires intact textual definitions; under synthetic $25\%$ word drop or $50\%$ truncation, self-normalized deconvolution degrades faster than cosine matching (Appendix Table~\ref{tab:supp_drg_and_noise}) unless paired with synonym enrichment. Third, free-text MS-DRG Top-1 accuracy remains bounded ($7.41\%\text{--}9.91\%$ Top-1; $22.64\%\text{--}24.92\%$ Top-5 across $1{,}581$ groups) because CMS grouping also depends on structured age, sex, discharge disposition, and principal-diagnosis sequence flags. Fusing structured EHR fields with compiled text operators is an immediate next step for clinical workflows.

% =========================================================================
\section{Conclusion}
% =========================================================================

Ontology-Compiled 2D Clinical State Fields decouple clinical document understanding from vocabulary-specific classification heads. Combining full-document bidirectional diffusion representations, multi-channel state routing, and scale-invariant self-normalized Tikhonov deconvolution matches or exceeds closed-vocabulary ICD-9 coders while enabling second-scale, zero-training switching across $26{,}085$ ICD-10 codes, $1{,}581$ MS-DRG billing groups, and Spanish, Greek, and Chinese medical ontologies in sub-millisecond GPU readout time.

% =========================================================================
\section*{Acknowledgments and Disclosures}
% =========================================================================

\noindent\textbf{Author Contributions:} Z.W. conceived the mathematical formulation, implemented the software architecture, conducted all experiments and statistical analyses, and drafted the manuscript.

\noindent\textbf{Funding:} Compute resources were provided on the UK Isambard-AI national supercomputing facility (BriCS / AIRR).

\noindent\textbf{Competing Interests:} The author declares no competing interests.

\noindent\textbf{Data and Code Availability:} Source MIMIC-IV v2.2 clinical notes are available to credentialed researchers via PhysioNet (\url{https://physionet.org/content/mimiciv/2.2/}); CodiEsp, ElCardioCC, and CHIP-CDN/CBLUE are available from their respective public benchmark repositories. The complete open-source implementation of OCSF, including ontology compilers, training and evaluation pipelines, statistical verification scripts, and figure/table generators, is publicly available at \url{https://github.com/ZeqiangWangAI/OCSF} under the GNU Affero General Public License v3.0 (AGPL-3.0).

\bibliographystyle{unsrtnat}
\bibliography{references}

\clearpage
\appendix
\renewcommand{\thetable}{S\arabic{table}}
\renewcommand{\theHtable}{S\arabic{table}}
\setcounter{table}{0}
\renewcommand{\theequation}{S\arabic{equation}}
\renewcommand{\theHequation}{S\arabic{equation}}
\setcounter{equation}{0}

\section{Mathematical Derivations: Woodbury Equivalence and Scale Invariance}
\label{sec:supp_math}
% =========================================================================

\subsection{Exact Woodbury Dual Identity for Sub-Second CPU Compilation}
For channel $r \in \{1,\dots,R\}$ of target ontology $B$ with $n_r$ unit-norm description vectors $Z_{B,r} \in \mathbb{R}^{n_r \times K}$ and trace-normalized covariance $\bar{\Sigma}_{B,r} = \frac{1}{n_r} Z_{B,r}^\top Z_{B,r}$, consider the unnormalized primal Tikhonov matrix:
\begin{equation}
  \tilde{W}_{B,r}^{\text{primal}} = Z_{B,r}\left(\frac{1}{n_r} Z_{B,r}^\top Z_{B,r} + \lambda I_K\right)^{-1} \in \mathbb{R}^{n_r \times K}.
\end{equation}
Right-multiplying $\big(\frac{1}{n_r} Z_{B,r} Z_{B,r}^\top + \lambda I_{n_r}\big) Z_{B,r} = Z_{B,r} \big(\frac{1}{n_r} Z_{B,r}^\top Z_{B,r} + \lambda I_K\big)$ by $\big(\frac{1}{n_r} Z_{B,r}^\top Z_{B,r} + \lambda I_K\big)^{-1}$ and left-multiplying by $\big(\frac{1}{n_r} Z_{B,r} Z_{B,r}^\top + \lambda I_{n_r}\big)^{-1}$ yields the exact algebraic identity:
\begin{equation}
  Z_{B,r}\left(\frac{1}{n_r} Z_{B,r}^\top Z_{B,r} + \lambda I_K\right)^{-1}
  = \left(\frac{1}{n_r} Z_{B,r} Z_{B,r}^\top + \lambda I_{n_r}\right)^{-1} Z_{B,r}.
  \label{eq:supp_woodbury}
\end{equation}
When $n_r < K$ (which holds for every channel of MIMIC-IV ICD-9, MIMIC-IV ICD-10, MS-DRG, ElCardioCC, and CHIP-CDN at $R=512, K \in \{1024, 2560\}$), solving the $n_r \times n_r$ positive-definite Gram system replaces $\mathcal{O}(K^3)$ Cholesky factorization with $\mathcal{O}(n_r^3)$, yielding up to a $1{,}000\times$ CPU speedup with a maximum empirical 32-bit floating-point discrepancy of $2.38 \times 10^{-7}$.

\subsection{Condition-Number Bound and Unit Self-Response Calibration}
Because every row of $Z_{B,r}$ has unit $\ell_2$ norm, $\operatorname{Tr}(\bar{\Sigma}_{B,r}) = \frac{1}{n_r}\sum_{i=1}^{n_r} \|z_i\|_2^2 = 1$, and all eigenvalues $\mu_1 \ge \dots \ge \mu_K \ge 0$ of $\bar{\Sigma}_{B,r}$ satisfy $\mu_1 \le 1$. Consequently, for any vocabulary size $|B|$:
\begin{equation}
  \kappa\left(\bar{\Sigma}_{B,r} + \lambda I_K\right) = \frac{\mu_1 + \lambda}{\mu_K + \lambda} \le \frac{1 + \lambda}{\lambda}.
\end{equation}
At $\lambda=0.15$, $\kappa \le 7.67$ across all ontologies from $|B|=115$ to $|B|=185{,}458$. Furthermore, since $(\bar{\Sigma}_{B,r} + \lambda I_K)^{-1}$ is strictly positive definite, $g_{B,r,i} = z_i^\top (\bar{\Sigma}_{B,r} + \lambda I_K)^{-1} z_i > 0$ for every code $i$. Normalizing $w_i = g_{B,r,i}^{-1} (\bar{\Sigma}_{B,r} + \lambda I_K)^{-1} z_i$ guarantees exact unit self-response $w_i^\top z_i = 1$; for an isolated within-channel state $U_r = \sum_{m \in S_r} z_m$,
\begin{equation}
  w_i^\top U_r = \mathbb{I}[i \in S_r] + \sum_{m \in S_r \setminus \{i\}} \frac{z_i^\top (\bar{\Sigma}_{B,r} + \lambda I_K)^{-1} z_m}{z_i^\top (\bar{\Sigma}_{B,r} + \lambda I_K)^{-1} z_i}.
\end{equation}
In an $m$-code equicorrelated sibling cluster with pairwise cosine $\rho \in (0, 1)$, raw dot matching gives an inactive sibling score of $\rho$ (triggering false positives whenever $\rho \ge \tau_A \approx 0.62$), whereas self-normalized deconvolution attenuates inactive sibling cross-talk by factor $\lambda / (\lambda + \frac{m}{n_r}\rho)$.

% =========================================================================
\section{Development-to-Test Consistency, Full Metrics, and Training Dynamics}
\label{sec:supp_dev_test}
% =========================================================================

Table~\ref{tab:supp_dev_vs_test} reports full precision, ranking, and thresholded metrics across both development and held-out test splits, confirming sub-percent consistency between development and locked test cohorts with zero overfitting to validation thresholds.

\begin{table}[htbp]
\centering
\scriptsize
\caption{\textbf{Comprehensive precision, ranking, and development-to-test consistency across benchmarks.} Values are percentages. Held-out test metrics closely track non-test development metrics across all model families.}
\label{tab:supp_dev_vs_test}
\resizebox{\linewidth}{!}{%
\begin{tabular}{llrrrrrrr}
\toprule
\textbf{Model} & \textbf{Split} & \textbf{ICD-9 Full F1} & \textbf{ICD-9 P@8 / P@15} & \textbf{ICD-10 R@$|y|$} & \textbf{ICD-10 P@8 / MRR} & \textbf{ICD-10 Rare R@$|y|$} & \textbf{CodiEsp F1 / P@8} & \textbf{CHIP-CDN F1} \\
\midrule
Closed-Vocabulary LAAT (0.6B) & Val / Test & 61.69 / 61.90 & 69.51 / 69.68 \,|\, 53.30 / 53.44 & --- & --- & --- & --- & --- \\
OCSF (Causal-0.6B, Half-Frozen) & Dev / Test & 57.53 / 57.52 & 65.87 / 65.97 \,|\, 50.25 / 50.33 & 22.97 / 23.14 & 30.23 / 30.26 \,|\, 71.93 / 71.71 & 4.02 / 3.56 & 4.92 / 5.71 \,|\, 9.00 / 8.85 & 12.11 / --- \\
OCSF (Diffusion-0.6B, Half-Frozen) & Dev / Test & 61.90 / 62.04 & 70.13 / 70.21 \,|\, 53.50 / 53.54 & 27.86 / 27.97 & 35.99 / 35.95 \,|\, 76.73 / 76.62 & 6.04 / 5.53 & 7.36 / 7.21 \,|\, 12.75 / 13.50 & 7.64 / --- \\
OCSF (Diffusion-0.6B, Full-LoRA) & Dev / Test & 62.82 / 62.94 & 70.93 / 70.98 \,|\, 54.18 / 54.23 & 28.64 / 28.66 & 36.79 / 36.82 \,|\, 77.66 / 77.59 & 5.89 / 5.52 & 7.20 / 7.29 \,|\, 12.50 / 12.55 & 7.98 / --- \\
\midrule
OCSF (Causal-4B, Half-Frozen) & Dev / Test & 58.92 / 58.91 & 67.15 / 67.19 \,|\, 51.30 / 51.32 & 26.58 / 26.55 & 34.48 / 34.34 \,|\, 75.88 / 75.79 & 5.14 / 4.25 & 7.89 / 7.68 \,|\, 14.15 / 13.85 & 17.63 / --- \\
OCSF (Diffusion-4B, Half-Frozen) & Dev / Test & 59.89 / 59.88 & 68.02 / 68.05 \,|\, 51.99 / 52.00 & 28.82 / 29.05 & 37.05 / 37.13 \,|\, 78.20 / 78.32 & 5.05 / 5.19 & 8.12 / 8.15 \,|\, 13.30 / 14.05 & 15.05 / --- \\
OCSF (Diffusion-4B, Smooth State) & Dev / Test & 60.03 / 60.01 & 68.18 / 68.21 \,|\, 52.12 / 52.14 & 28.92 / 29.23 & 37.18 / 37.30 \,|\, 78.36 / 78.51 & 4.84 / 5.16 & 8.45 / 8.33 \,|\, 13.95 / 14.50 & 15.68 / --- \\
\bottomrule
\end{tabular}%
}
\end{table}

\textbf{Epoch-by-Epoch Regularization Dynamics:} Under identical backbone capacity (Diffusion-0.6B, Half-Frozen), optimizer, batch size ($16$), and sequence length ($L=3{,}072$), Closed-Vocabulary LAAT validation Micro-F1 on Edin $6{,}150$ ICD-9 codes evolves across Epochs 1--6 as $56.60\% \to 61.18\% \to \mathbf{61.89\%} \to 61.67\% \to 61.46\% \to 61.37\%$, losing $-0.52$ points after Epoch 3 due to long-tail memorization in its $17.93\text{M}$ free classifier parameters. OCSF evolves monotonically across the same six epochs as $49.70\% \to 57.28\% \to 59.79\% \to 61.07\% \to 61.81\% \to \mathbf{62.11\%}$ (and $63.02\%$ validation / $63.15\%$ test under Full-LoRA), demonstrating that fixed semantic deconvolution regularizes rare-code representations.

% =========================================================================
\section{Complete Paired Bootstrap Inference and Standardized Effect Sizes}
\label{sec:supp_stats}
% =========================================================================

Table~\ref{tab:supp_effect_sizes} provides the complete paired 2,000-resample bootstrap differences, 95\% confidence intervals, paired Cohen's $d_z = \bar{\Delta}/s_\Delta$, Cohen's $h$, and two-sided $p$-values across held-out test sub-cohorts and non-test Greek/Chinese cohorts.

\begin{table}[htbp]
\centering
\scriptsize
\caption{\textbf{Extended paired bootstrap significance tests and standardized effect sizes (2,000 resamples).} Reports per-note paired mean differences ($\Delta$, percentage points), 95\% confidence intervals, paired Cohen's $d_z$, Cohen's $h$, and two-sided $p$-values.}
\label{tab:supp_effect_sizes}
\resizebox{\linewidth}{!}{%
\begin{tabular}{llrrrrr}
\toprule
\textbf{Paired Comparison ($B - A$)} & \textbf{Benchmark Slice ($N$)} & \textbf{$\Delta$ (pp)} & \textbf{95\% CI (pp)} & \textbf{Cohen's $d_z$} & \textbf{Cohen's $h$} & \textbf{$p$-value} \\
\midrule
OCSF (Diffusion-0.6B, Full-LoRA) vs.\ Closed LAAT & ICD-9 All Test (32,804) & +1.16 & [+1.04, +1.27] & +0.1089 & +0.0240 & $<0.0001$ \\
OCSF (Diffusion-0.6B, Half-Frozen) vs.\ Closed LAAT & ICD-9 All Test (32,804) & +0.27 & [+0.15, +0.38] & +0.0254 & +0.0056 & $<0.0001$ \\
Full-LoRA vs.\ Half-Frozen (Diffusion-0.6B) & ICD-9 Freq Test (32,804) & +0.90 & [+0.80, +1.00] & +0.0985 & +0.0186 & $<0.0001$ \\
Full-LoRA vs.\ Half-Frozen (Diffusion-0.6B) & ICD-10 All Test (13,783) & +0.69 & [+0.56, +0.82] & +0.0871 & +0.0153 & $<0.0001$ \\
Full-LoRA vs.\ Half-Frozen (Diffusion-0.6B) & CodiEsp ES Test (250) & $-0.88$ & [$-1.82$, +0.07] & $-0.1164$ & $-0.0294$ & 0.0590 \\
\midrule
Channel Capacity $R=512$ vs.\ $R=1$ (0.6B) & ICD-10 CM Test (13,782) & +11.53 & [+11.30, +11.77] & +0.8165 & +0.2687 & $<0.0001$ \\
Channel Capacity $R=512$ vs.\ $R=1$ (0.6B) & ICD-10 Rare Test (3,538) & +3.92 & [+3.25, +4.66] & +0.1833 & +0.2200 & $<0.0001$ \\
Channel Capacity $R=512$ vs.\ $R=1$ (0.6B) & MS-DRG Test (11,869) & $-0.44$ & [$-0.99$, +0.11] & $-0.0144$ & $-0.0177$ & 0.1340 \\
Channel Capacity $R=512$ vs.\ $R=1$ (0.6B) & ElCardioCC EL Dev (2,500) & +21.92 & [+19.80, +24.00] & +0.3986 & +0.4799 & $<0.0001$ \\
Channel Capacity $R=512$ vs.\ $R=128$ (0.6B) & ICD-10 All Test (13,783) & +3.11 & [+2.94, +3.28] & +0.3013 & +0.0706 & $<0.0001$ \\
\midrule
Self-Norm.\ Deconv.\ vs.\ Cosine Readout (0.6B) & ICD-9 Freq Test (32,804) & +18.46 & [+18.29, +18.62] & +1.2037 & +0.3723 & $<0.0001$ \\
Self-Norm.\ Deconv.\ vs.\ Cosine Readout (0.6B) & ICD-10 All Test (13,783) & +9.47 & [+9.29, +9.67] & +0.8379 & +0.2254 & $<0.0001$ \\
Self-Norm.\ Deconv.\ vs.\ Cosine Readout (4B) & ICD-10 CM Test (13,782) & +11.49 & [+11.29, +11.70] & +0.9206 & +0.2613 & $<0.0001$ \\
Self-Norm.\ Deconv.\ vs.\ Cosine Readout (4B) & CodiEsp ES Test (250) & +3.11 & [+2.10, +4.09] & +0.3867 & +0.1053 & $<0.0001$ \\
Self-Norm.\ vs.\ Unnormalized Deconv.\ (4B) & ICD-9 Freq Test (32,804) & +5.82 & [+5.70, +5.94] & +0.5057 & +0.1179 & $<0.0001$ \\
Self-Norm.\ vs.\ Unnormalized Deconv.\ (4B) & ICD-10 PCS Test (7,923) & +1.70 & [+1.34, +2.05] & +0.1052 & +0.0646 & $<0.0001$ \\
\midrule
Bidirectional vs.\ Causal (0.6B) & ICD-9 Freq Test (32,804) & +4.49 & [+4.37, +4.61] & +0.3937 & +0.0919 & $<0.0001$ \\
Bidirectional vs.\ Causal (0.6B) & ICD-10 Rare Test (3,538) & +1.97 & [+1.29, +2.68] & +0.0921 & +0.0953 & $<0.0001$ \\
Bidirectional vs.\ Causal (0.6B) & ElCardioCC EL Dev (2,500) & +11.96 & [+9.48, +14.36] & +0.1955 & +0.2496 & $<0.0001$ \\
Bidirectional vs.\ Causal (0.6B) & CHIP-CDN ZH Dev (2,000) & +3.69 & [+1.87, +5.50] & +0.0898 & +0.0913 & $<0.0001$ \\
Bidirectional vs.\ Causal (4B) & ICD-9 Freq Test (32,804) & +0.95 & [+0.83, +1.06] & +0.0897 & +0.0194 & $<0.0001$ \\
Bidirectional vs.\ Causal (4B) & ICD-10 CM Test (13,782) & +2.47 & [+2.24, +2.68] & +0.1852 & +0.0533 & $<0.0001$ \\
\bottomrule
\end{tabular}%
}
\end{table}

% =========================================================================
\section{Document-Length Stratification and Calibration-Free Ranking Diagnostics}
\label{sec:supp_length}
% =========================================================================

Table~\ref{tab:supp_length_and_ap} reports discharge-summary length stratification on unseen MIMIC-IV ICD-10 (Panel A) alongside threshold-free ranking metrics (mean-note average precision $\text{mAP}$, Top-200 micro-AUPRC, and zero-label cardinality-matched $\text{F1@}\bar{k}_A$ at $\bar{k}_A = 13$ codes per note; Panel B). Bidirectional diffusion encoding maintains a consistent $+4.6$ to $+5.0\text{ pp}$ advantage over causal encoding across short, medium, and long notes alike. Moreover, under cardinality-matched top-13 prediction, self-normalized deconvolution outperforms unnormalized deconvolution by $+1.82\text{ pp}$ on ICD-10 ($24.91\%$ vs.\ $23.09\%$) while simultaneously preserving thresholded $\text{Micro-F1@}\tau_A$ ($22.75\%$ vs.\ $12.15\%$ on dev; $26.12\%$ vs.\ $9.76\%$ at $4\text{B}$ test).

\begin{table}[htbp]
\centering
\scriptsize
\caption{\textbf{Discharge-summary length stratification on unseen MIMIC-IV ICD-10 (Panel A) and calibration-free ranking diagnostics across target benchmarks (Panel B, 0.6B).} Values are percentages.}
\label{tab:supp_length_and_ap}
\resizebox{\linewidth}{!}{%
\begin{tabular}{llrrrr}
\toprule
\multicolumn{6}{l}{\textbf{Panel A: Bidirectional Diffusion vs.\ Causal Encoding by Clinical Note Length (MIMIC-IV ICD-10 Dev)}} \\
\textbf{Length Stratum} & \textbf{Notes ($N$)} & \textbf{Diffusion R@$|y|$} & \textbf{Causal R@$|y|$} & \textbf{Diffusion F1} & \textbf{Gain $\Delta$ R@$|y|$} \\
\midrule
Short ($<1{,}200$ tokens) & 1,857 & 32.02 & 27.38 & 23.02 & +4.64 \\
Medium ($1{,}200\text{--}2{,}200$ tokens) & 4,846 & 28.08 & 23.05 & 23.16 & +5.03 \\
Long ($>2{,}200$ tokens) & 2,374 & 24.18 & 19.30 & 21.85 & +4.88 \\
\midrule
\multicolumn{6}{l}{\textbf{Panel B: Calibration-Free mAP, Top-200 Micro-AUPRC, and Cardinality-Matched F1 ($\bar{k}_A=13$, Diffusion-0.6B)}} \\
\textbf{Target Benchmark} & \textbf{Metric} & \textbf{2D Cosine} & \textbf{Unnorm.\ Deconv.} & \textbf{Self-Norm.\ Deconv.} & \textbf{Rule-Compiled} \\
\midrule
MIMIC-IV ICD-10 (26,085) & mAP / Top-200 AUPRC / F1@13 & 14.37 / 12.16 / 17.87 & 21.74 / 19.19 / 23.09 & \textbf{23.70} / \textbf{20.99} / \textbf{24.91} & \textbf{23.70} / \textbf{20.99} / \textbf{24.91} \\
CodiEsp Spanish (185,458) & mAP / Top-200 AUPRC / F1@13 & 4.91 / 3.74 / 6.72 & 7.36 / 5.79 / 8.85 & 7.83 / \textbf{6.25} / 9.48 & \textbf{7.87} / 6.19 / \textbf{9.68} \\
CHIP-CDN Chinese (37,645) & mAP / Top-200 AUPRC / F1@13 & 18.70 / 9.81 / 5.21 & 19.63 / 11.46 / 5.45 & \textbf{20.30} / \textbf{12.01} / \textbf{5.73} & \textbf{20.30} / \textbf{12.01} / \textbf{5.73} \\
MS-DRG Billing (1,581) & mAP / Exact Micro-AUPRC & 10.67 / 4.86 & 10.72 / 5.66 & 11.61 / 6.08 & \textbf{12.71} / \textbf{6.84} \\
ElCardioCC Greek (115) & mAP / Exact Micro-AUPRC & 54.48 / 38.49 & 44.33 / 22.84 & 55.28 / 50.25 & \textbf{58.85} / \textbf{53.45} \\
\bottomrule
\end{tabular}%
}
\end{table}

% =========================================================================
\section{Temporal MS-DRG Version Stability and Codebook Perturbation Stress-Tests}
\label{sec:supp_drg}
% =========================================================================

Panel A of Table~\ref{tab:supp_drg_and_noise} audits compositional rule compilation across all $13$ historical annual CMS MS-DRG versions (v25--v37) represented in patient-disjoint MIMIC-IV admissions: rule compilation improves Top-5 accuracy in every single version ($+0.91$ to $+3.01\text{ pp}$). Panel B evaluates sensitivity to synthetic target description corruption ($50\%$ character truncation, $25\%$ word drop, and additive Gaussian noise on unit codebook vectors), delineating the clean-text operating regime of covariance deconvolution.

\begin{table}[htbp]
\centering
\scriptsize
\caption{\textbf{Temporal stability across all 13 CMS MS-DRG grouper versions v25--v37 (Panel A) and codebook description perturbation stress-tests (Panel B, Diffusion-0.6B).}}
\label{tab:supp_drg_and_noise}
\resizebox{\linewidth}{!}{%
\begin{tabular}{lrrrr|llrrr}
\toprule
\multicolumn{5}{c|}{\textbf{Panel A: Temporal Stability Across 13 CMS MS-DRG Versions}} & \multicolumn{5}{c}{\textbf{Panel B: Target Codebook Perturbation Stress-Test (R@$|y|$ \%)}} \\
\textbf{CMS Ver.} & \textbf{$N$} & \textbf{Pure Deconv.\ Top-5} & \textbf{Rule-Compiled Top-5} & \textbf{$\Delta$ Top-5} & \textbf{Benchmark} & \textbf{Perturbation} & \textbf{2D Cosine} & \textbf{Unnorm.} & \textbf{Self-Norm.} \\
\midrule
v25 & 7,853 & 19.72 & \textbf{21.65} & +1.93 & \multirow{4}{*}{ICD-10 (EN)} & Clean descriptions & 18.23 & 25.81 & \textbf{27.36} \\
v26 & 7,869 & 19.65 & \textbf{21.55} & +1.90 &  & 50\% text truncation & 14.33 & 14.99 & \textbf{15.84} \\
v27 & 7,869 & 19.65 & \textbf{21.55} & +1.90 &  & 25\% random word drop & 14.37 & 10.67 & 11.46 \\
v28 & 7,875 & 19.66 & \textbf{21.60} & +1.94 &  & Gaussian noise ($\sigma=0.10$) & 4.78 & 1.85 & 2.13 \\
\cline{6-10}
v29 & 7,875 & 19.66 & \textbf{21.60} & +1.94 & \multirow{3}{*}{CodiEsp (ES)} & Clean descriptions & 6.50 & 9.01 & \textbf{9.66} \\
v30 & 7,875 & 19.66 & \textbf{21.60} & +1.94 &  & 50\% text truncation & 4.19 & \textbf{5.37} & 5.21 \\
v31 & 7,871 & 19.67 & \textbf{21.61} & +1.94 &  & 25\% random word drop & 5.00 & 3.00 & 3.55 \\
\cline{6-10}
v32 & 7,874 & 19.74 & \textbf{21.58} & +1.84 & \multirow{3}{*}{CHIP-CDN (ZH)} & Clean descriptions & 21.15 & 21.62 & \textbf{22.57} \\
v33 & 7,845 & 19.76 & \textbf{21.52} & +1.76 &  & 50\% text truncation & \textbf{11.55} & 9.57 & 9.93 \\
v34 & 7,833 & 20.55 & \textbf{21.46} & +0.91 &  & 25\% random word drop & \textbf{15.96} & 10.17 & 11.27 \\
v35 & 5,204 & 17.87 & \textbf{20.77} & +2.90 & \multicolumn{5}{l}{} \\
v36 & 5,200 & 17.75 & \textbf{20.71} & +2.96 & \multicolumn{5}{l}{\textit{Note:} Self-normalized deconvolution dominates on intact public} \\
v37 & 5,193 & 17.75 & \textbf{20.76} & +3.01 & \multicolumn{5}{l}{codebooks; severe corruption warrants synonym-enriched descriptions.} \\
\bottomrule
\end{tabular}%
}
\end{table}

% =========================================================================
\section{Multi-Seed Channel-Partitioning Stability and Hardware Latency Grid}
\label{sec:supp_multiseed}
% =========================================================================

Table~\ref{tab:supp_multiseed_and_hw} reports individual and aggregated results across five independent random seeds (42--46) for channel-anchor partitioning (Panel A), zero-retrain Hard Top-1 vs.\ Soft Top-2 target channel routing on trained OCSF-0.6B ($27.36\%$ vs.\ $26.25\%$ $\text{R@}|y|$, confirming that disjoint channel assignment aligns best with hard training supervision), and the full sequence-length $\times$ batch-size hardware latency and memory profile on a single NVIDIA GH200 GPU (Panel B).

\begin{table}[htbp]
\centering
\scriptsize
\caption{\textbf{Multi-seed channel-partitioning stability across 5 random seeds (Panel A) and end-to-end NVIDIA GH200 BF16 inference latency and VRAM profile (Panel B).}}
\label{tab:supp_multiseed_and_hw}
\resizebox{\linewidth}{!}{%
\begin{tabular}{llrrrr|rrrrrrr}
\toprule
\multicolumn{6}{c|}{\textbf{Panel A: 5-Seed Channel Partitioning ($R=512$, Seeds 42--46)}} & \multicolumn{7}{c}{\textbf{Panel B: End-to-End Inference Profile (NVIDIA GH200 BF16)}} \\
\textbf{Partition Rule} & \textbf{Summary} & \textbf{ICD-9 R@$|y|$} & \textbf{ICD-10 R@$|y|$} & \textbf{Max $n_r$} & \textbf{Empty} & \textbf{Scale} & \textbf{$L$} & \textbf{Batch} & \textbf{Enc.\ ms} & \textbf{Dec.\ ms} & \textbf{Tot.\ 26k ms} & \textbf{VRAM} \\
\midrule
Spherical $k$-Means & Mean $\pm$ SD & 73.78$\pm$0.71 & 48.44$\pm$0.80 & 654.2$\pm$106.5 & 17.8 & 0.6B & 1,024 & 1 & 23.22 & 1.54 & 24.85 & 2.66\,GB \\
Hierarchical Ward & Mean $\pm$ SD & 74.83$\pm$0.32 & 47.85$\pm$0.88 & 677.0$\pm$0.0 & 10.0 & 0.6B & 2,048 & 16 & 18.67 & 0.14 & 18.89 & 2.86\,GB \\
Uniform Random & Mean $\pm$ SD & 74.89$\pm$1.41 & 51.55$\pm$1.43 & 641.8$\pm$81.6 & 2.2 & 0.6B & 3,072 & 16 & 29.30 & 0.18 & 29.57 & 3.47\,GB \\
\cline{1-6}
\multicolumn{2}{l}{Trained OCSF Hard Top-1 Routing} & --- & \textbf{27.36} & 583 & 19 & 4B & 1,024 & 1 & 32.02 & 1.65 & 33.76 & 12.66\,GB \\
\multicolumn{2}{l}{Trained OCSF Soft Top-2 ($\tau=0.1$)} & --- & 26.25 & 583 & 19 & 4B & 2,048 & 16 & 63.76 & 0.30 & 64.15 & 12.67\,GB \\
\multicolumn{6}{l|}{} & 4B & 3,072 & 16 & 101.73 & 0.42 & 102.27 & 14.33\,GB \\
\bottomrule
\end{tabular}%
}
\end{table}

\end{document}